%% file: samplepaper.tex
\documentclass[runningheads]{llncs}
\usepackage[T1]{fontenc}
\usepackage{graphicx}
\usepackage{cite} 
\usepackage[colorlinks,allcolors=blue]{hyperref} 

\usepackage{multirow} 
\usepackage{makecell} 
\usepackage{subfig} 
\usepackage{booktabs} 
\usepackage{tabularx}  

\usepackage{amssymb} 
\usepackage{mathtools} 

\DeclarePairedDelimiter\abs{\lvert}{\rvert}
\DeclarePairedDelimiter\floor{\lfloor}{\rfloor} 

\DeclarePairedDelimiterX{\infdivx}[2]{(}{)}{%
  #1\;\delimsize\|\;#2%
}

\usepackage[size=footnotesize]{todonotes}

\begin{document}
\title{Towards Fairness and Privacy: A Novel Data Pre-processing Optimization Framework for Non-binary Protected Attributes\thanks{This work was supported by the Federal Ministry of Education and Research (BMBF) under Grand No.~16DHB4020.}
}
\titlerunning{Towards Fairness and Privacy: A Novel Data Pre-processing Framework}
%
\author{Manh Khoi Duong\orcidID{0000-0002-4653-7685} \and \\
Stefan Conrad\orcidID{0000-0003-2788-3854}}
\authorrunning{M. K. Duong \and S. Conrad}
%
\institute{Heinrich Heine University, Universit\"atsstra\ss{}e 1, 40225 D\"usseldorf, Germany
\email{\{manh.khoi.duong, stefan.conrad\}@hhu.de}}

%
\maketitle              
\begin{abstract}
	The reason behind the unfair outcomes of AI is often rooted in biased datasets.
	Therefore, this work presents a framework for addressing fairness by debiasing datasets containing a (non-)binary protected attribute.
	The framework proposes a combinatorial optimization problem where heuristics such as genetic algorithms can be used to solve for the stated fairness objectives.
	The framework addresses this by finding a data subset that minimizes a certain discrimination measure. Depending on a user-defined setting, the framework enables different use cases, such as data removal, the addition of synthetic data, or exclusive use of synthetic data. The exclusive use of synthetic data in particular enhances the framework's ability to preserve privacy while optimizing for fairness.
	In a comprehensive evaluation, we demonstrate that under our framework, genetic algorithms can effectively yield fairer datasets compared to the original data.
	In contrast to prior work, the framework exhibits a high degree of flexibility as it is metric- and task-agnostic, can be applied to both binary or non-binary protected attributes, and demonstrates efficient runtime.
\end{abstract}
\input{text/introduction}
\input{text/preliminaries}
\input{text/problem}
\input{text/method}
\input{text/evaluation}
\input{text/conclusion}


%
%
%
\bibliographystyle{splncs04}
\bibliography{references}
\end{document}

%% file: text/introduction.tex
\section{Introduction}
\begin{figure}[htbp]
  \centering
  \includegraphics[width=0.98\textwidth]{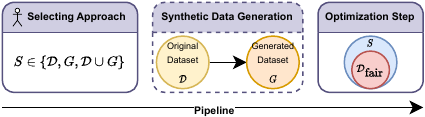}
  \caption{The pipeline consists of three steps: (1) The user sets the sample set $S$ and other settings, including the objective, discrimination measure, and protected attribute; (2) Synthetic data is generated if needed; (3) A solver optimizes the fairness objective to yield a biased-reduced subset $\mathcal{D}_\text{fair}$ from the user-selected set $S$. If $S = G$ was chosen, the user obtains a bias-reduced synthetic dataset that does not leak privacy-related information.}
  \label{fig:pipeline}
\end{figure}

Machine learning has become an increasingly important tool for decision-making in various applications, ranging from income~\cite{ron1996_adult} to recidivism prediction~\cite{larson_angwin_mattu_kirchner_2016}.
However, the use of these models can perpetuate existing biases in the data and result in unfair treatment of certain demographic groups.
One of the key concerns in the development of fair machine learning models is the prevention of discrimination regarding protected attributes such as race, gender, and religion.

While most of the existing literature focuses on classification problems where the protected attribute is binary~\cite{friedrich2023FairDiffusion, mehrabi2021survey, dunkelau2019fairness, caton2020fairness, barocas-hardt-narayanan2019book, mindiffprost, zemel2013learning, duonggen2023}, the real world presents a more complex scenario where the protected attribute can consist of more than two social groups, making it non-binary.
While works that discuss and deal with non-binary protected attributes exist,
and we do not neglect their existence~\cite{liobait2017MeasuringDI, noisypacelis21a, kamani2022fairpca}, we view it as a necessity to contribute further
to this field by providing a flexible framework that
accommodates various fairness notions and applications, including data privacy,
to strive for the employment of responsible artificial intelligence in practice.

Since bias is rooted in data, we introduce an optimization
framework that pre-processes data to mitigate discrimination.
In the context of fairness, pre-processing ensures
the generation of a fair, debiased dataset.
We address the challenges associated with non-binary protected attributes by
deriving appropriate discrimination measures.
To prevent discrimination, we formulate a combinatorial optimization problem
to identify a subset from a given sample dataset that minimizes a specific discrimination measure, as depicted in Fig.~\ref{fig:pipeline}.
Depending on the provided sample dataset, which may also include synthetically generated data,
the framework allows for the removal of such data points or the inclusion of synthetic ones to achieve equitable outcomes.
By using generated data, we can utilize our method in
applications where data privacy is a concern.
Since the discrimination objective is stated as a black box,
heuristics, which do not assess the analytical expression of the discrimination measure
during optimization, are needed to solve our stated problem.
Our formulation makes the framework \emph{fairness-agnostic}, allowing it to be used
to pursue any fairness objective.

The experimentation was carried out on the Adult~\cite{ron1996_adult}, Bank~\cite{moro2014bank}, and COMPAS~\cite{larson_angwin_mattu_kirchner_2016} datasets,
all known to exhibit discrimination.
We compared the discrimination of the datasets before and after pre-processing them with different heuristics on various discrimination measures.
The results show that genetic algorithms~\cite{holland1975adaptationgenetic} were most effective in reducing discrimination
for non-binary protected attributes.
To summarize, the primary contributions of this paper are:

\begin{itemize}
	\item We present an optimization framework that renders different approaches for yielding fair data. The approaches include removing, adding generated data, or solely using generated data.
	\item We underscore the framework's ability to handle cases where data privacy is a significant concern.
	\item Our methodology is designed to handle a protected attribute that can be non-binary, offering broader applicability.
	\item We carry out an extensive evaluation of the proposed techniques on three biased datasets. The evaluation focuses on their effectiveness in reducing discrimination and their runtimes.
	\item We publish our implementation at \url{https://github.com/mkduong-ai/fairdo} as a documented Python package and distribute it over PyPI.
\end{itemize}

\section{Related Work}
Recently, related works have equivalently formulated subset selection problems
to achieve fairness goals~\cite{tang2023beyond, duonggen2023}.
While in the work of Tang et al.~\cite{tang2023beyond}, a distribution is generated that
represents the selection probability of each feasible set to maximize the
global utility on average, our work aims to return a definite subset.
To achieve fairness according to any defined criteria,
our formulation treats discrimination measures as black boxes.
These measures can encompass both group and individual fairness notions,
distinguishing our work from that of
Tang et al.~\cite{tang2023beyond}, whose framework is limited to
group fairness.

Previous studies have also utilized synthetic data to address
fairness and privacy concerns~\cite{duonggen2023, liu2023generating}.
Both of these studies employed heuristics similar to our approach.
In particular, Liu et al.\cite{liu2023generating} specialized
on generating synthetic data using a genetic algorithm to
satisfy specific privacy definitions\cite{dwork2006differential, bun2016concentrated}.
While our framework does not generate privacy-preserving data specifically,
it utilizes synthetic data, which can be generated with such methods.
Similarly to our work, Duong et al.~\cite{duonggen2023} leveraged synthetic data
by introducing a sampling-based heuristic for
selecting a subset of such data points to minimize discrimination.
Our work generalizes the work of Duong et al.\cite{duong2023generate} as
their approach can be viewed as a special case of ours.
Additionally, our formulation offers greater flexibility compared to the approach of
Duong et al.\cite{duong2023generate}, as it
allows for any heuristic to tackle the task and
is also not limited to binary protected attributes.

%% file: text/preliminaries.tex
\section{Measuring Discrimination}\label{section:measuring}
In this section, we introduce the notation used to derive discrimination measures for assessing dataset fairness: A \emph{data point} or \emph{sample} is represented as a triple $(x, y, z)$, where $x \in X$ is the \emph{feature}, $y \in Y$ is the ground truth \emph{label} indicating favorable or unfavorable outcomes, and $z \in Z$ is the \emph{protected attribute}, which is used to differentiate between groups. The sets $X, Y, Z$ typically hold numeric values and are defined as $X = \mathbb{R}^d$, $Y=\{0, 1\}$, and $Z = \{1, 2, \ldots, k\}$ with $k \geq 2$.
For instance, in the context of predicting personal attributes, we can use $X$ to represent numeric values that encode particular aspects of a person.
$Y$ typically describes the positive or negative outcome that we aim to predict for the person.
$Z$ can denote any protected attribute, such as race, which can be used to identify the person as Caucasian, Afro-American, Latin American, or Asian.
We assume that $z$ is not included as a feature in $x$. To be able to differentiate between groups, $k \geq 2$ must hold. If $k > 2$, the protected attribute $Z$ is said to be non-binary.
Following the definition, a \emph{dataset}, denoted as $\mathcal{D} = \{d_i\}_{i=1}^n$, consists of data points,
where a single sample is defined as $d_i = (x_i, y_i, z_i)$.
Machine learning models are trained using these datasets to predict the target variable $y$ based on the input variables $x$ and $z$.
Finally, we denote a discrimination measure with $\psi\colon \mathbb{D} \to [0, 1]$,
where $\mathbb{D}$ is the set of all datasets.

In the following, $x, y, z$ are noted as random variables that can take on specific values.

\subsection{Absolute Measures}
To deal with non-binary groups, Žliobaitė~\cite{liobait2017MeasuringDI} suggested in her work to compare groups pairwise.
For this, she presented three possible ways which are comparing each group with another,
one against the rest for each group, and all groups against the unprivileged group.
The author further discussed options to aggregate the results.
Although Žliobaitė~\cite{liobait2017MeasuringDI} stated textually how to measure
discrimination for more than two groups, we express them mathematically in this work.
To treat groups equally without presuming which group is unprivileged
and to get the full picture, we choose to make use of comparing each group with another.
We first introduce the common fairness notion \emph{statistical parity}~\cite{kamishima2012fairness, zemel2013learning}, which demands equal positive outcomes for different groups in $Z = \{1, 2, \ldots, k\}$.
It is usually defined for binary groups, but we present the non-binary cases~\cite{liobait2017MeasuringDI}.

\begin{definition}[Statistical parity]\label{def:statistical-parity}
Demanding that each of the $k$ groups have the same probability
of receiving the favorable outcome is statistical parity, i.e.,
\begin{align*}
	& P(y=1 \mid z=1) = \ldots = P(y=1 \mid z=k)\\
	\iff \quad & P(y=1 \mid z=i) = P(y=1 \mid z=j) \quad \forall i, j \in Z.
\end{align*}
\end{definition}
As the group size $k$ grows, the satisfaction of statistical parity becomes less probable.
Because of this, the equality constraints are treated softly by deriving
differences between the groups. Consequently, smaller differences imply more equality.
For binary groups, the difference is often referred to as statistical disparity (SDP)~\cite{dunkelau2019fairness}.

\begin{definition}[Sum of absolute statistical disparities]\label{def:sumspabs}
Let there be $k$ groups, then the sum of absolute statistical disparities is calculated as follows~\cite{liobait2017MeasuringDI}:
\begin{align*}
	\psi_\text{SDP-sum}(\mathcal{D}) &= \sum_{\substack{i, j \in Z \\ i \neq j, j \geq i}} \abs{P(y=1 \mid z=i) - P(y=1 \mid z=j)}\\
	&= \sum_{i=1}^{k} \sum_{j=i+1}^{k} \abs{P(y=1 \mid z=i) - P(y=1 \mid z=j)}.
\end{align*}
Because the total number of comparisons is $\frac{k(k-1)}{2}$~\cite{liobait2017MeasuringDI},
the average discrimination between all groups becomes:
\begin{align*}
	\psi_\text{SDP-avg}(\mathcal{D}) = \frac{2}{k(k-1)} \cdot \sum_{i=1}^{k} \sum_{j=i+1}^{k} \abs{P(y=1 \mid z=i) -
	P(y=1 \mid z=j)}.
\end{align*}
\end{definition}

\begin{definition}[Maximal absolute statistical disparity]\label{def:maxspabs}
Maximal absolute statistical disparity measures the absolute statistical disparity between all pairs $i, j \in Z$ and returns the maximum value.
Specifically, it is given by:
\begin{align*}
	\psi_\text{SDP-max}(\mathcal{D}) = \max_{i, j \in Z} \abs{P(y=1 \mid z=i) -
	P(y=1 \mid z=j)}.
\end{align*}
\end{definition}

Žliobaitė~\cite{liobait2017MeasuringDI}, after consulting with legal experts,
recommends using the maximum function to aggregate disparities,
though the choice depends on the ethical context of the specific use case.
Discrimination measures can be seen as social welfare functions.
Minimizing the sum of absolute statistical disparities
is analogous to the utilitarian viewpoint~\cite{mill1863utilitarianism},
which aims to maximize the general utility of the population.
If one decides to care for the least well-off group, then minimizing the maximal absolute statistical disparity corresponds to the Rawlsian social welfare~\cite{rawls1971theory}.

%% file: text/problem.tex
\section{Optimization Framework}\label{section:problemformulation}
Inspired by related works that identify unfair data samples~\cite{verma2021removing, kamiran2012data}, we propose a method to remove such samples for fairness.
The task is formulated as a combinatorial problem where the aim is to determine a subset $\mathcal{D}_\text{fair}$ of a given set $S$ such that the discrimination of the subset $\psi(\mathcal{D}_\text{fair})$ is minimal, as shown in Fig.~\ref{fig:pipeline}.
Depending on the application, set $S$ can be the original data $\mathcal{D}$,
a synthetic set $G$ with the same distribution as $\mathcal{D}$,
or their union $\mathcal{D} \cup G$.

\subsection{Problem Formulation}
To state the problem mathematically, let note $S = \{s_1, s_2, \ldots, s_{\tilde{n}}\}$ and further introduce a binary vector $b$ with the same length as $S$, i.e.,
$b = (b_1, b_2, \ldots, b_{\tilde{n}})$.
To define the combinatorial optimization problem, each entry $b_i$ in $b$ is examined whether it is 1 ($b_i = 1$), in which case the corresponding sample $s_i$ in $S$ is included in the subset $\mathcal{D}_\text{fair}$. Therefore, the fair set is defined with
\begin{equation}
	\mathcal{D}_\text{fair} = \{s_i \in S \mid b_i = 1, i=1 \ldots \tilde{n}\}.
\end{equation}
The objective $f\colon {0, 1}^{\tilde{n}} \to [0, 1]$ can then be expressed by:
\begin{align}
	& f_{S, \psi}(b) = \psi(\mathcal{D}_\text{fair}) \nonumber \\
	\iff \quad & f_{S, \psi}(b) = \psi(\{s_i \in S \mid b_i = 1, i=1 \ldots \tilde{n}\}),
\end{align}
where $f_{S, \psi}$ is defined as the discrimination of a subset $\mathcal{D}_\text{fair}$ of the given set $S$ and $\psi$ evaluates the level of discrimination on $\mathcal{D}_\text{fair}$.
Note that the decision variable is $b$, for which $\mathcal{D}_\text{fair}$ can be obtained.
The subindices $S$ and $\psi$ of $f_{S, \psi}$ can be seen as settings for the objective. 
Ignoring the subindices, we write out the combinatorial optimization problem as follows:
\begin{align}\label{eq:objective}	
	\min_{b} & \quad f(b)  \\ \nonumber
	\textrm{subject to} & \quad b_i \in \{0,1\} \quad \forall i = 1,\ldots,\tilde{n}. \nonumber
\end{align}

\noindent Because the set of feasible subsets $\mathcal{P}(S)$ grows exponentially regarding
the cardinality of $S$, we employ heuristics to solve our stated problem.

In the following subsections, we discuss different and useful settings of $S$ that serve different purposes with their corresponding advantages and disadvantages.

\subsection{Removing Samples ($S = \mathcal{D}$)}
By setting $S = \mathcal{D}$, it is intended to determine data points in the training set that can be removed to prevent discrimination.
Intuitively, having an overexposure of certain types of samples that fulfill stereotypes can result in a discriminatory dataset. 
In such situations, the most practical step is to remove the affected samples.

However, this method is not recommended if the given dataset is small.
Likewise, some could argue that minorities can be easily removed by this method. Luckily, this can be prevented by choosing the right discrimination measure.

\subsection{Employing Only Synthetic Data ($S = G$)}\label{section:syntheticdata}
To employ synthetic data, this method relies on a statistical model.
The statistical model is used to learn
the distribution of the original data $P(\mathcal{D})$.
By doing so, synthetic samples $G$ can be drawn from the learned distribution
$G \sim P(\mathcal{D})$.

Relying solely on synthetic data is particularly important in use cases where data privacy and protection are major concerns and the use of real data is prohibited.
Of course, synthetic data is not necessarily disjoint from the original dataset and can therefore be a privacy breach itself.
For tabular and smaller datasets, this can be naively mitigated by removing such privacy breaching points from the synthetic data by setting $S = G \setminus \mathcal{D}$.
Other ways include populating differential privacy techniques in the data generation process~\cite{dwork2006differential, jordon2019pate, abay2019privacy, liu2023generating}.

When generally using synthetic data, one cannot easily ensure that the corresponding label of the features is correct.
Training machine learning models on synthetic data can therefore lead to higher error rates when predicting on real data.
Despite the distribution of the synthetic data following
the distribution of the real dataset, it depends heavily on the method used
when it comes to generating qualitative, faithful data.

\subsection{Merging Real and Synthetic Data ($S = \mathcal{D} \cup G$)}
Another approach to generate a non-discriminatory dataset is to merge the original dataset $\mathcal{D}$ with synthetic data $G$ that has been generated with a statistical model as described in Section~\ref{section:syntheticdata}.
By combining the two sets $S = \mathcal{D} \cup G$, it is possible to increase the size of the resulting dataset while avoiding over-representation of discriminatory samples.

One advantage of this method is that it can improve the quality of the data by utilizing both the real $\mathcal{D}$ and synthetic data $G$. The resulting dataset can be larger and more diverse, which can lead to greater robustness when training machine learning models. If the dataset is too small to apply removal techniques ($S = \mathcal{D}$) or relying solely on synthetic data ($S = G$) appears unreliable, merging the two sets may be a viable option.

However, this method is not without its limitations and comes with disadvantages when generally using synthetic data, e.g., quality and faithfulness.
Different from the method described in Section~\ref{section:syntheticdata}, this method is not applicable for purposes with privacy concerns as samples from the real data are not omitted.

\subsection{Adding Synthetic Data}\label{section:addsynthetic}
A different approach that requires a new formulation of the objective is to
include synthetic data points without deleting any samples from the real data.
As well, a set of generated data points $G$ must be given,
and the research question is which of the generated points can lead to a fairer distribution when including them in the original dataset.
The possible use case for this problem is to fine-tune machine learning models that
have already learned from an unfair dataset.
This is mostly useful for large machine learning models where resources are scarce to retrain the whole model. Following the preceding notation, the fair dataset becomes:
\begin{equation}
	\mathcal{D}_\text{fair}^\text{add} = \mathcal{D} \cup \{s_i \in S \mid b_i = 1, i=1 \ldots \tilde{n}\}
\end{equation}
and we express the corresponding objective $f_{S, \psi}^{\text{add}}$ by:
\begin{align}
	& f_{ S, \psi}^\text{add}(b) = \psi(\mathcal{D}_\text{fair}^\text{add}) \nonumber \\
	\iff \quad & f_{S, \psi}^\text{add}(b) = \psi(\mathcal{D} \cup \{s_i \in S \mid b_i = 1, i=1 \ldots \tilde{n}\}),
\end{align}
where $S$ is set to $G$ to achieve the described approach. Certainly, $S$ can also be set to $\mathcal{D}$ or any other set operation on $\mathcal{D}$ with $G$.
Although such settings are possible, they do not serve any meaningful purposes.
However, one could argue that setting $S = \mathcal{D}$ can act as a reweighing method.
Still, we argue against facilitating duplicates in a dataset with intent,
as no additional information is provided.

As seen, our framework offers many advantages due to its versatility and therefore
potential use in a broad range of applications.
By choosing the appropriate objective function, discrimination measure, and sample set, the formulation is tailored to the specific intent and use case.
Because the formulation is agnostic to the solver,
it can serve multiple purposes without modifying solvers.

%% file: text/method.tex
\section{Heuristics}\label{section:methodology}
This section presents heuristics that specifically solve combinatorial optimization problems. These include: a baseline method that returns the original dataset,
a simple random heuristic, and genetic algorithms with different operators.

\begin{enumerate}
	\item \textbf{Original}: Uses the original data by returning a vector of ones $b = \mathbf{1}_{\tilde{n}}$.
	\item \textbf{Random Heuristic}: Generates a user-defined number of random vectors,
        with each entry having a 50\% chance of being zero or one, and then returns the best solution.
    \item \textbf{Genetic Algorithm (GA)}: The workflow of GAs~\cite{introductionevolution} involves generating an initial population of candidate solutions and then repeatedly performing \emph{selection}, \emph{crossover}, and \emph{mutation} operations over several generations. In our implementation, the GA terminates earlier if improved solutions were not
    found within 50 generations. Following operators were used in our experimentation~\cite{goldberg1996genetic}:
    \begin{itemize}
    		\item Selection: \emph{Elitist}, \emph{Tournament}, \emph{Roulette Wheel} (see~\cite{goldberg1996genetic} for more details)
    		\item Crossover: \emph{Uniform} (each entry of the offspring has the same probability of either inheriting the entry from the first or second parent)
    		\item Mutation: \emph{Bit Flip} (flips a fixed amount of random bits for each vector, that is $\floor{p_m \cdot \tilde{n}}$, where $p_m \in [0, 1]$ is the mutation rate)
    \end{itemize}
\end{enumerate}

%% file: text/evaluation.tex
\begin{table}[tb]
\caption{Overview of Datasets}
\centering
\label{table:dataset_comparison}
\begin{tabularx}{\textwidth}{lrrlXX}
\hline
\textbf{Dataset} & \textbf{Entries} & \textbf{Cols.} & \textbf{Label} & \textbf{Protected Attribute} & \textbf{Description}\\
\hline
Adult~\cite{ron1996_adult} & 32\,561	& 22	& Income & Race: White, Black, Asian-Pacific-Islander, American-Indian-Eskimo, Other & Indicates individuals earning over \$50,000 annually \\ \hline
Bank~\cite{moro2014bank}	& 41\,188	& 53 	& \begin{tabular}[t]{@{}l@{}} Term deposit\\ subscription\end{tabular} & Job: Admin, Blue-Collar, Technician, Services, Management, Retired, Entrepreneur, Self-Employed, Housemaid, Unemployed, Student, Unknown & Shows whether the client has subscribed to a term deposit.\\ \hline
COMPAS~\cite{larson_angwin_mattu_kirchner_2016}	& 7\,214	& 8	& \begin{tabular}[t]{@{}l@{}} 2-year\\ recidivism\end{tabular} & Race: African-American, Caucasian, Hispanic, Other, Asian, Native American & Displays individuals that were rearrested for a new crime within 2 years after initial arrest.\\
\hline
\end{tabularx}
\end{table}

\section{Evaluation}
In our evaluation, we conducted multiple experiments to address the following research questions:
\begin{itemize}
	\item \textbf{RQ1}\enspace How do the heuristics perform in making the datasets fairer?
	\item \textbf{RQ2}\enspace How does runtime vary among the heuristics?
	\item \textbf{RQ3}\enspace How stable are the results across the runs?
	\item \textbf{RQ4}\enspace Is there a clear winner? If yes, which method is recommended for practical use?
\end{itemize}

To answer these research questions, we specifically designed
experiments for the Adult~\cite{ron1996_adult}, Bank~\cite{moro2014bank},
and COMPAS~\cite{larson_angwin_mattu_kirchner_2016} datasets.
Both the Adult and COMPAS datasets include race as a non-binary protected attribute,
whereas the Bank dataset utilizes the job as a non-binary protected attribute.
All datasets were prepared and cleansed in the same manner: Categorical features were
one-hot encoded, with the exception of the protected attribute and the label.
Additionally, rows containing missing values were excluded from all datasets.
Table~\ref{table:dataset_comparison} shows details about the datasets used in our experiments after the preparation and cleansing steps.

Following the dataset preparation, we executed two distinct experiments.
The first experiment (Section~\ref{section:hypertuning}) was dedicated to hyperparameter tuning of the GAs,
adjusting both population sizes and the number of generations to pinpoint optimal configurations.
Armed with these optimal settings, our second experiment (Section~\ref{section:comparing}) focused on
comparing different selection operators within GAs (\textbf{RQ1}).
Our aim was to determine which operator yielded the best performance.
This experiment included comparisons to several baseline methods,
one of which simply returned the original data.
By expanding our evaluation to multiple discrimination measures in this phase,
we can comprehensively assess the effectiveness of GAs
in reducing discrimination in datasets.

The experimental methodology involves the application of heuristics
to produce a binary mask, which yields fair data.
We then measure the discrimination of the resulting dataset.
To ensure stability in our findings (\textbf{RQ3}),
each experiment was repeated 15 times.
We additionally recorded the runtime of each trial to tackle \textbf{RQ2}.
Depending on the experiment, we employed suitable heuristics that aim to solve
each objective with the associated discrimination measure,
as listed in Table~\ref{table:expsettings}.
For instance, each heuristic either optimizes $f_{S, \psi}$ or $f_{ S, \psi}^\text{add}$
with varying settings of $S$ and $\psi$ as given in the table.
In order to perform experiments with synthetic data,
we generated data that has the same size as the original dataset,
i.e., $|G| = |\mathcal{D}|$.
The statistical model used to generate synthetic data is
Gaussian copula~\cite{patki2016_copula} which is fast and performs well on tabular data.
For privacy-sensitive use cases, we advise utilizing privacy-preserving techniques~\cite{dwork2006differential, jordon2019pate, abay2019privacy, liu2023generating}.
All experiments were conducted on an Intel(R) Xeon(R) Gold 5120 processor
clocking at 2.20GHz.

\begin{table}[tb]
\caption{Configuration details of heuristics, objectives, and discrimination measures for each experiment.}
\centering
\begin{tabular}{l|l|l|l}
\hline
Experiment     & Heuristics                                                                                             & Objectives ($f$, $S$)                                        & Disc. Measures ($\psi$)           \\ \hline
Hyperparam. & GA                                                                                                     & Remove, Merge, Add                                                         & Sum SDP \\ \hline
Comparison     & \begin{tabular}[t]{@{}l@{}}Original, Random,\\ GA (Elitist, Tournament,\\ Roulette Wheel)\end{tabular} & Remove, Merge, Add & \begin{tabular}[t]{@{}l@{}}Sum SDP, Max SDP\end{tabular} \\ \hline
\end{tabular}
\label{table:expsettings}
\end{table}

\subsection{Hyperparameter Tuning}\label{section:hypertuning}
For the genetic algorithm, we performed hyperparameter tuning, exploring various population sizes [20, 50, 100, 200] and generations [50, 100, 200, 500], all using tournament selection, uniform crossover, and bit flip mutation at a rate of 5\%. These configurations are described in Section~\ref{section:methodology}.
We evaluated the algorithm on three distinct objectives and set $\psi_\text{SDP-sum}$
as the discrimination measure.

\subsubsection{Discrimination}
\begin{figure}[tb]
\centering
\begin{tabular}{ccc}
\multicolumn{1}{c}{Adult|Remove} & \multicolumn{1}{c}{Merge} & \multicolumn{1}{c}{Add} \\
\includegraphics[width=0.32\textwidth]{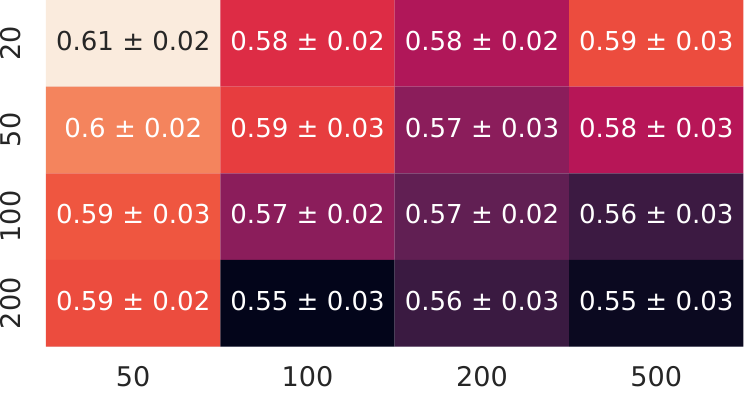} & 
\includegraphics[width=0.32\textwidth]{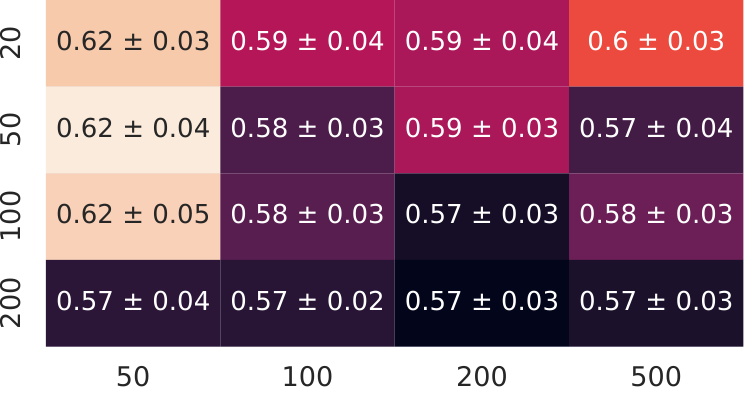} & 
\includegraphics[width=0.32\textwidth]{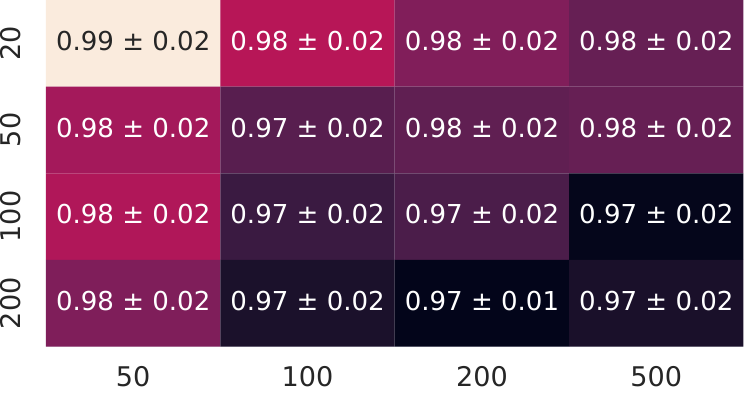} \\
Bank & & \\
\includegraphics[width=0.32\textwidth]{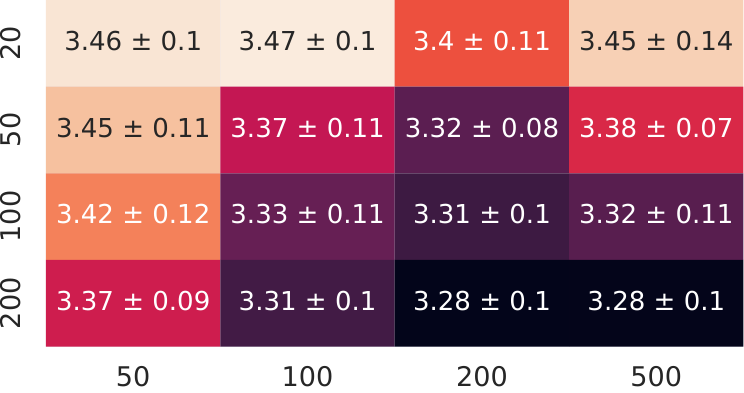} & 
\includegraphics[width=0.32\textwidth]{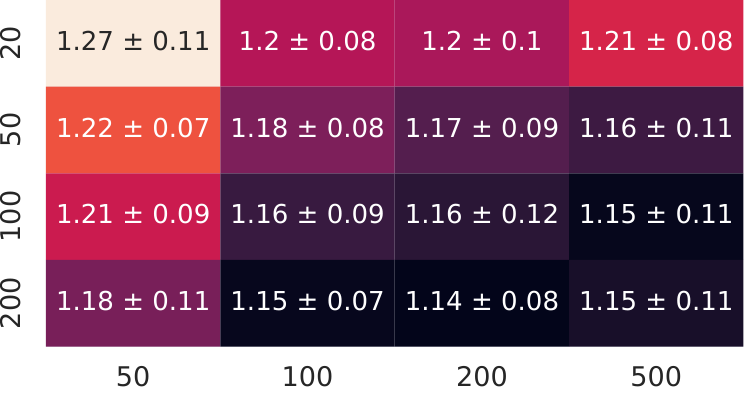} & 
\includegraphics[width=0.32\textwidth]{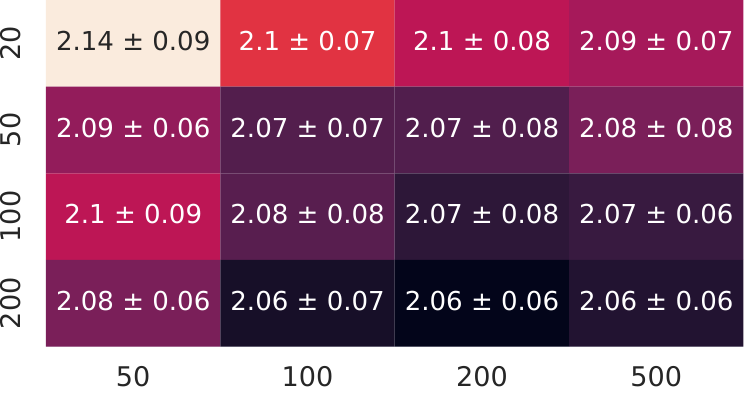} \\
COMPAS & & \\
\includegraphics[width=0.32\textwidth]{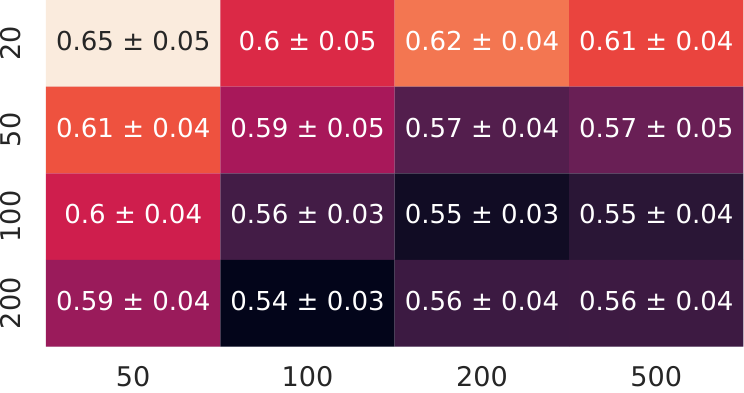} & 
\includegraphics[width=0.32\textwidth]{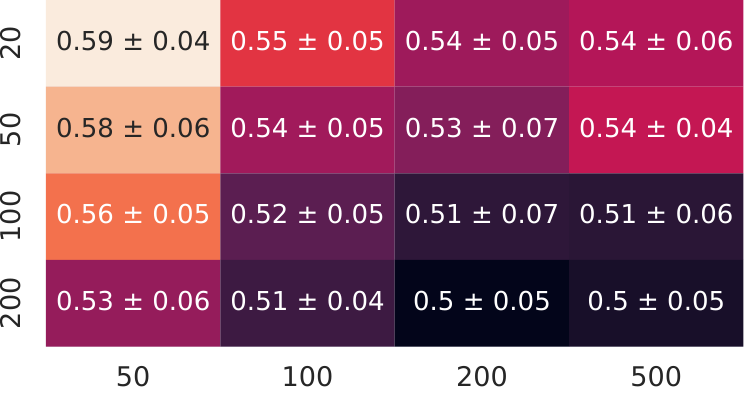} & 
\includegraphics[width=0.32\textwidth]{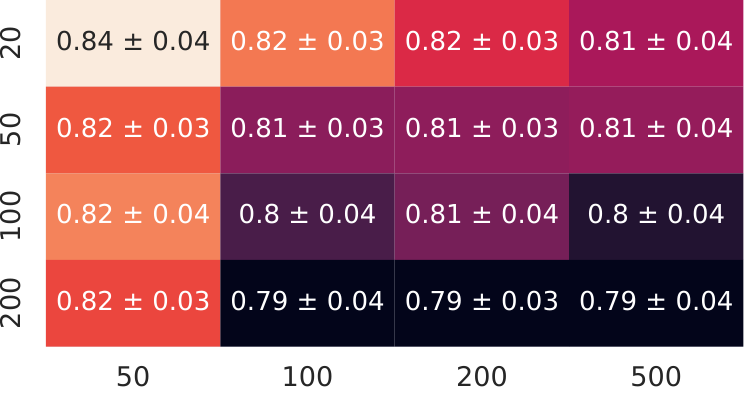} \\
\end{tabular}
\caption{Heatmaps showing discrimination scores ($\psi_\text{SDP-sum}$) after pre-processing with genetic algorithms using different population sizes (y-axis) and generations (x-axis). Rows depict the results of Adult, Bank, and COMPAS datasets, while columns represent the objectives.}
\label{fig:hyp_performance}
\end{figure}

As seen in Fig.~\ref{fig:hyp_performance},
the heatmaps display the average discrimination (including the standard deviation) of
GAs solving various objectives on different datasets.
Each heatmap shows hyperparameters that were set for the experimentation.
Across the different objectives and datasets, there is a consistent trend indicating
that utilizing larger populations combined with a higher number of generations
typically results in less discrimination.
This is particularly evident when contrasting scenarios with a population size of
20 and 50 generations, which, on average,
have discrimination scores higher by 0.1.
However, the improvements in discrimination plateau beyond certain thresholds.
Specifically, once the number of generations surpasses 200 or when the population size
exceeds 100,
there is no significant further decrease in discrimination observable.

\subsubsection{Runtime}
\begin{figure}[tb]
\centering
\subfloat[Remove]{%
\includegraphics[width=0.33\textwidth]{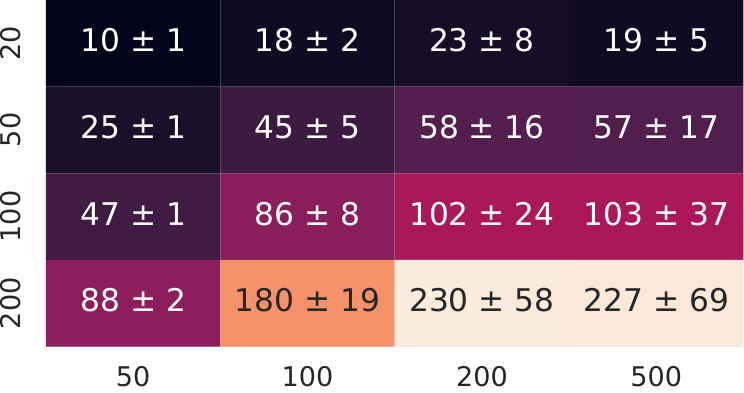}%
} \hfill
\subfloat[Merge]{%
\includegraphics[width=0.33\textwidth]{%
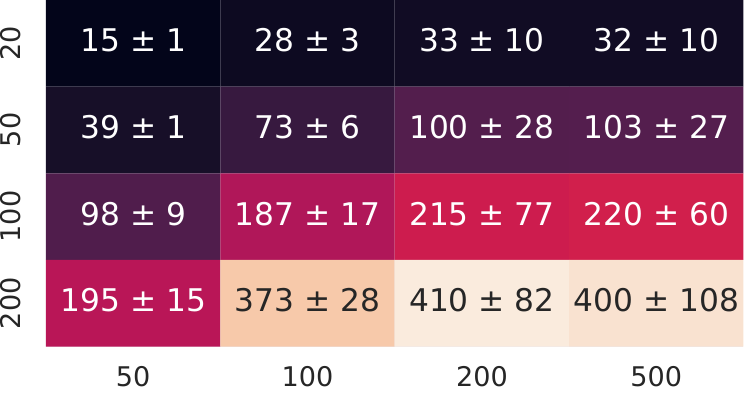}%
} \hfill
\subfloat[Add]{%
\includegraphics[width=0.33\textwidth]{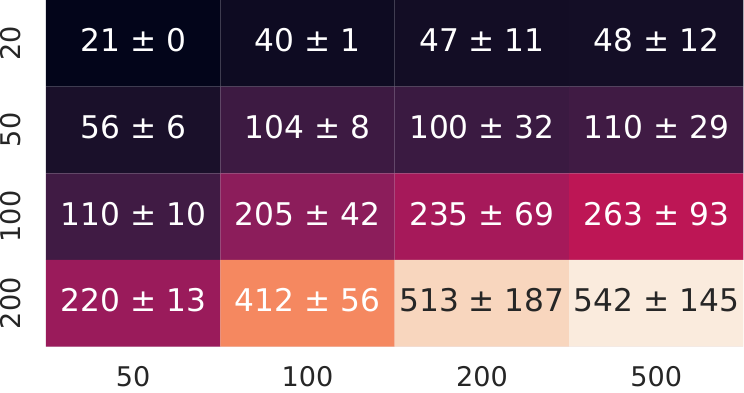}%
} \hfill
\caption{Heatmaps showing runtimes in seconds for the Bank dataset after pre-processing with genetic algorithms using different population sizes (y-axis) and generations (x-axis).}
\label{fig:hyp_time}
\end{figure}

For brevity reasons, we display the runtimes solely for the Bank dataset in Fig.~\ref{fig:hyp_time},
given its larger size and the similarity of the results across other datasets.
The outcome of this analysis pointed towards an optimal setting of a population size
of 100 combined with 500 generations.
Under our specifications, executing the GA with these settings 
takes, on average, between 1.5 and 4.5 minutes.
While increasing the population size further did not show significant
improvements in reducing the bias in the datasets,
it proved to be more efficient in terms of the runtime.

\subsection{Comparing Heuristics}\label{section:comparing}
After determining that a population size of 100 with 500 generations offered optimal results
w.r.t. discrimination and time,
this configuration was maintained for all subsequent experiments.
Here, three GAs were compared, each differing by their selection operator:
elitist, tournament, and roulette wheel selection.
All GAs were set with uniform crossover and bit flip mutation at a rate of 5\% to
perform the experiments.
Additionally, we established both the original dataset and the random heuristic as baselines.

\subsubsection{Discrimination}
\begin{table}[tb]
\caption{Displayed are the mean discrimination scores, accompanied by standard deviations, from 15 runs.
The heuristics were evaluated across multiple objectives using varying discrimination measures on the Adult, Bank, and COMPAS datasets. Best results are marked bold.}
\label{table:performance}
\centering
\resizebox{\textwidth}{!}{%
\begin{tabular}{llllllll}
\hline
 &  & \multicolumn{3}{l}{Sum SDP} & \multicolumn{3}{l}{Max SDP} \\
 &  & Adult & Bank & COMPAS & Adult & Bank & COMPAS \\
Objective & Method &  &  &  &  &  &  \\
\hline
\multirow[t]{5}{*}{Add} & 1. Original & 1.07 ± 0.02 & 1.83 ± 0.09 & 1.17 ± 0.06 & 0.23 ± 0.00 & 0.09 ± 0.00 & 0.17 ± 0.01 \\
 & 2. Random & 1.03 ± 0.02 & 2.27 ± 0.07 & 0.94 ± 0.03 & 0.21 ± 0.00 & 0.11 ± 0.00 & 0.15 ± 0.01 \\
 & 3. Elitist & \textbf{0.82 ± 0.02} & \textbf{1.54 ± 0.06} & \textbf{0.59 ± 0.03} & \textbf{0.16 ± 0.00} & \textbf{0.07 ± 0.00} & \textbf{0.10 ± 0.00} \\
 & 4. Tournament & 0.97 ± 0.02 & 2.06 ± 0.06 & 0.80 ± 0.03 & 0.20 ± 0.00 & 0.10 ± 0.00 & 0.13 ± 0.00 \\
 & 5. Roulette & 1.03 ± 0.02 & 2.31 ± 0.08 & 0.94 ± 0.05 & 0.21 ± 0.00 & 0.11 ± 0.00 & 0.15 ± 0.01 \\
\cline{1-8}
\multirow[t]{5}{*}{Merge} & 1. Original & 1.07 ± 0.02 & 1.83 ± 0.09 & 1.17 ± 0.06 & 0.23 ± 0.00 & 0.09 ± 0.00 & 0.17 ± 0.01 \\
 & 2. Random & 0.80 ± 0.03 & 1.46 ± 0.09 & 0.76 ± 0.08 & 0.16 ± 0.01 & 0.07 ± 0.00 & 0.12 ± 0.01 \\
 & 3. Elitist & \textbf{0.21 ± 0.04} & \textbf{0.42 ± 0.07} & \textbf{0.11 ± 0.05} & \textbf{0.04 ± 0.01} & \textbf{0.02 ± 0.00} & \textbf{0.01 ± 0.00} \\
 & 4. Tournament & 0.58 ± 0.04 & 1.17 ± 0.09 & 0.51 ± 0.04 & 0.11 ± 0.01 & 0.05 ± 0.00 & 0.09 ± 0.01 \\
 & 5. Roulette & 0.85 ± 0.05 & 1.49 ± 0.09 & 0.79 ± 0.09 & 0.16 ± 0.01 & 0.07 ± 0.00 & 0.12 ± 0.01 \\
\cline{1-8}
\multirow[t]{5}{*}{Remove} & 1. Original & 0.97 ± 0.00 & 4.81 ± 0.00 & 1.89 ± 0.00 & 0.17 ± 0.00 & 0.25 ± 0.00 & 0.27 ± 0.00 \\
 & 2. Random & 0.71 ± 0.02 & 4.07 ± 0.07 & 0.72 ± 0.03 & 0.12 ± 0.00 & 0.19 ± 0.00 & 0.12 ± 0.01 \\
 & 3. Elitist & \textbf{0.25 ± 0.02} & \textbf{1.41 ± 0.12} & \textbf{0.20 ± 0.07} & \textbf{0.05 ± 0.00} & \textbf{0.07 ± 0.01} & \textbf{0.01 ± 0.00} \\
 & 4. Tournament & 0.57 ± 0.02 & 3.29 ± 0.08 & 0.56 ± 0.04 & 0.11 ± 0.00 & 0.15 ± 0.01 & 0.09 ± 0.01 \\
 & 5. Roulette & 0.75 ± 0.03 & 4.15 ± 0.10 & 0.75 ± 0.08 & 0.13 ± 0.00 & 0.20 ± 0.01 & 0.12 ± 0.01 \\
\cline{1-8}
\end{tabular}
}
\end{table}

Table~\ref{table:performance} presents the discrimination results of our experiments.
It is evident that all tested algorithms are stable, as reflected by the low standard deviations (\textbf{RQ3}).
All heuristics were able to reduce the discrimination available in the datasets in most cases.
Elitist selection consistently outperformed other methods,
offering notable improvements in fairness compared to the original datasets (\textbf{RQ1}).
We emphasize that the measures handle non-binary attributes,
providing flexibility in targeting various fairness goals.
Further, by the range of discrimination measures utilized,
our methodology can aim for diverse fairness goals,
be it the enhancement of the utilitarian social welfare ($\psi_\text{SDP-sum}$) or
Rawlsian social welfare ($\psi_\text{SDP-max}$), as evidenced.
An interesting observation from our study is the varied discrimination levels based on the specific measure used, as seen in the Bank dataset, where
its discrimination is either highest or lowest when compared with other datasets.
This is due to the higher number of groups,
leading to more group comparisons that affect the overall discrimination score.
When examining the objectives, removing both the
synthetic and original data tends to outperform others.
This observation is particularly evident in the Merge objective.
Given the consistent performance of the elitist selection in our tests, we strongly recommend its use for those aiming to achieve the best fairness outcomes (\textbf{RQ4}).

\subsubsection{Runtime}
\begin{table}[tb]
\caption{Mean runtimes in seconds of different methods solving different objectives with varying discrimination measures on the Adult, Bank, and COMPAS datasets.}
\label{table:runtimes}
\resizebox{\textwidth}{!}{%
\begin{tabular}{llllllll}
\hline
 &  & \multicolumn{3}{l}{Sum SDP} & \multicolumn{3}{l}{Max SDP} \\
 &  & Adult & Bank & COMPAS & Adult & Bank & COMPAS \\
Objective & Method &  &  &  &  &  &  \\
\hline
\multirow[t]{5}{*}{Add} & 1. Original & \textbf{0 ± 0} & \textbf{0 ± 0} & \textbf{0 ± 0} & \textbf{0 ± 0} & \textbf{0 ± 0} & \textbf{0 ± 0} \\
 & 2. Random & 50 ± 1 & 107 ± 12 & 14 ± 0 & 51 ± 6 & 103 ± 7 & 13 ± 0 \\
 & 3. Elitist & 320 ± 105 & 605 ± 224 & 53 ± 21 & 334 ± 80 & 636 ± 179 & 79 ± 23 \\
 & 4. Tournament & 122 ± 38 & 209 ± 50 & 39 ± 17 & 119 ± 37 & 216 ± 74 & 34 ± 12 \\
 & 5. Roulette & 82 ± 26 & 131 ± 46 & 26 ± 9 & 82 ± 40 & 132 ± 48 & 26 ± 12 \\
\cline{1-8}
\multirow[t]{5}{*}{Merge} & 1. Original & \textbf{0 ± 0} & \textbf{0 ± 0} & \textbf{0 ± 0} & \textbf{0 ± 0} & \textbf{0 ± 0} & \textbf{0 ± 0} \\
 & 2. Random & 46 ± 3 & 67 ± 1 & 15 ± 2 & 44 ± 4 & 66 ± 1 & 15 ± 3 \\
 & 3. Elitist & 283 ± 103 & 359 ± 143 & 79 ± 25 & 286 ± 111 & 397 ± 161 & 75 ± 28 \\
 & 4. Tournament & 127 ± 39 & 185 ± 69 & 36 ± 11 & 131 ± 61 & 169 ± 51 & 44 ± 19 \\
 & 5. Roulette & 69 ± 21 & 127 ± 53 & 28 ± 9 & 83 ± 33 & 118 ± 31 & 29 ± 14 \\
\cline{1-8}
\multirow[t]{5}{*}{Remove} & 1. Original & \textbf{0 ± 0} & \textbf{0 ± 0} & \textbf{0 ± 0} & \textbf{0 ± 0} & \textbf{0 ± 0} & \textbf{0 ± 0} \\
 & 2. Random & 23 ± 1 & 44 ± 1 & 11 ± 0 & 22 ± 1 & 47 ± 11 & 11 ± 0 \\
 & 3. Elitist & 138 ± 66 & 281 ± 119 & 52 ± 18 & 176 ± 68 & 290 ± 80 & 50 ± 15 \\
 & 4. Tournament & 78 ± 13 & 119 ± 25 & 24 ± 9 & 73 ± 27 & 132 ± 46 & 25 ± 7 \\
 & 5. Roulette & 58 ± 27 & 72 ± 19 & 22 ± 8 & 52 ± 24 & 71 ± 24 & 18 ± 7 \\
\cline{1-8}
\end{tabular}
}
\end{table}

An analysis of the runtimes is presented in Table~\ref{table:runtimes}.
The original method consistently took 0 seconds (rounded) to finish.
At second comes the random method and lastly GAs.
The elitist operator took the longest, with runtimes approximately three times slower than the quickest operator, the roulette wheel. Tournament selection comes in between.
Most experiments were finished in 5 minutes or less, which is still very efficient.
Regarding the measures, the runtimes when optimizing $\psi_\text{SDP-max}$ appeared negligibly higher compared to $\psi_\text{SDP-sum}$, so it can be disregarded.
Generally, larger datasets yielded longer runtimes,
revealing a linear relationship between dataset size and runtime.
In addressing the research question posed in \textbf{RQ4},
it becomes evident that the elitist operator is superior among the tested methods.
Despite being the slowest method, it is still very efficient at reducing discrimination
on datasets consisting of up to 41\,188 samples, as seen in our experimentation.

%% file: text/conclusion.tex
\section{Conclusion}
We introduced a novel and flexible optimization framework to reduce discrimination
and preserve privacy in datasets. The framework accommodates various intents
such as data removal, synthetic data addition, and exclusive use of synthetic data for privacy reasons.
Notably, the objectives in our framework are designed to be independent of
specific discrimination measures, allowing users and stakeholders to address
any form of discrimination without modifying the solvers.

Due to the relatively sparse work existing on dealing with non-binary attributes,
particularly regarding established methods,
we tackled non-binary protected attributes in our experiments by deriving discrimination measures based on the work of Žliobaitė~\cite{liobait2017MeasuringDI} and showed that our framework allowed the effective and fast reduction of discrimination by
employing heuristics.

\section{Future Work and Discussion}
Future work could include extending the usability of this framework by deriving different discrimination measurements. Thus, handling multiple protected attributes as well as regression tasks can be done without modifying the general methodology. Additionally, formulating and integrating constraints into the objective function can also be done, which further enhances the responsibility of our approach. For instance, we could consider constraints such as group sizes and add penalties if samples of minorities get removed.

Although we aim for fairness and data privacy with our framework, it is still important to engage with diverse stakeholders to identify unintended consequences and address possible ethical implications.
Particularly, an extensive discussion and analysis of the used objective
and discrimination measure for a specific application should be done
to ensure that the data aligns with the desired fairness goals.
